\documentclass[11pt]{article}

\usepackage{iclr2026_conference}

\usepackage{times}
\usepackage{latexsym}
\usepackage[T1]{fontenc}
\usepackage[utf8]{inputenc}
\usepackage{microtype}
\usepackage{inconsolata}
\usepackage{graphicx}
\usepackage{booktabs}
\usepackage{multirow}
\usepackage{array}
\usepackage{xcolor}
\usepackage{colortbl}
\usepackage{amsmath,amssymb}
\usepackage{hyperref}
\usepackage{enumitem}
\usepackage[breakable,skins]{tcolorbox}
\usepackage[font=small,labelfont=bf,skip=4pt]{caption}

\usepackage{tikz}
\usetikzlibrary{arrows.meta,positioning,fit,backgrounds,calc,shapes.geometric,decorations.pathreplacing}
\usepackage{pgfplots}
\pgfplotsset{compat=1.18}

\definecolor{pdblue}{HTML}{1F77B4}
\definecolor{pdorange}{HTML}{D62728}
\definecolor{pdgreen}{HTML}{2CA02C}
\definecolor{pdgold}{HTML}{B8860B}
\definecolor{pdpurple}{HTML}{6A51A3}
\definecolor{pdgray}{HTML}{4B5563}
\definecolor{pdlight}{HTML}{F3F4F6}

\definecolor{anchGreen}{HTML}{EAF7EE}
\definecolor{anchYellow}{HTML}{FFF6D8}
\definecolor{anchRed}{HTML}{FDECEC}
\definecolor{anchRedStrong}{HTML}{F8D7DA}
\definecolor{anchGray}{HTML}{F2F4F7}
\definecolor{anchMuted}{HTML}{667085}

\newcommand{\prHigh}[1]{\cellcolor{anchGreen}\textbf{#1}}
\newcommand{\prMid}[1]{\cellcolor{anchYellow}#1}
\newcommand{\prLow}[1]{\cellcolor{anchRed}#1}
\newcommand{\prFloor}[1]{\cellcolor{anchRedStrong}\textbf{#1}}

\newcommand{\hall}{\textsc{Hall}}
\newcommand{\mis}{\textsc{Mis}}
\newcommand{\unc}{\textsc{Unc}}
\newcommand{\rec}{\textsc{Rec}}
\newcommand{\probe}{\textsc{ClaimProbe}}
\newcommand{\drwriter}{\textsc{ClaimWriter}}

\newcommand{\drbot}{\textsc{ClaimWriter-BU}}
\newcommand{\aiq}{AI-Q}
\newcommand{\edr}{EDR}
\newcommand{\edrlong}{Enterprise Deep Research}
\newcommand{\odr}{ODR}
\newcommand{\odrlong}{Open Deep Research}

\title{Redesigning and Auditing Deep Research Writing for Faithful Reports}
\iclrfinalcopy

\author{
Hiroaki Hayashi\thanks{Equal contribution.}\quad
Pranav Narayanan Venkit\footnotemark[1]\quad
Prafulla Kumar Choubey\footnotemark[1]\quad
Chien-Sheng Wu\\
Salesforce AI Research \\
\small{
  \textbf{Correspondence:}
  \href{mailto:hiroakihayashi@salesforce.com,pnarayananvenkit@salesforce.com,pchoubey@salesforce.com,wu.jason@salesforce.com}
  {[hiroakihayashi, pnarayananvenkit, pchoubey, wu.jason]@salesforce.com}
}
}

\begin{document}
\maketitle

\begin{abstract}
Rubric-based evaluations of deep-research (DR) systems often obscure fine-grained factual failures in generated reports. We introduce \probe{}, a claim-level audit that decomposes DR reports into claims and measures hallucination, misattribution, citation hygiene, and necessary-fact recall against retrieved evidence. Using \probe{}, we find that strong DR pipelines can omit key evidence and misattribute claims even when their rubric scores remain stable. We then propose \drwriter{}, a hierarchical claim-based writer that extracts source facts, maps them to a query-derived outline, and drafts each section from a source-linked claim representation. Across three prior DR frameworks, replacing only the report writer with \drwriter{} reduces hallucination by $2.6\times$--$4.5\times$ and improves necessary-fact recall by $1.2\times$--$1.7\times$, while largely preserving overall report quality. \drwriter{} also enables localized revision: when sources change, it propagates changed source facts into revised reports at the highest rate among update methods, while also being more cost-effective.

\end{abstract}

\noindent\textbf{Github:} \url{https://github.com/SalesforceAIResearch/claimwriter-deep-research}

\section{Introduction}
\label{sec:intro}

% Deep-research (DR) systems are agents that plan multi-step web search, gather sources, and produce long-form, citation-heavy reports \citep{venkit2025deeptrace}.
% Industry teams deploy them where careful synthesis is the deliverable: market analysis, regulatory due diligence, policy briefing, and scientific review.
% With potentially large impact on business efficiency and execution, leading DR systems are predominantly proprietary or closed-source~\citep{google2025geminideepresearch,openai2025deepresearch,deepresearchbench_leaderboard}, though open-source systems trail closely while also offering transparency that enables scientific research~\citep{choubey2026edr,nvidia_aiq,langchain2025odr}. %: \edrlong{} (\edr{}) \citep{choubey2026edr}, a recent enterprise pipeline with coverage-driven objectives; the \aiq{} Blueprint \citep{nvidia_aiq}, NVIDIA's LangGraph-based research agent; and \odrlong{} (\odr{}) \citep{langchain2025odr}, the leading open reference implementation.

Deep-research (DR) systems are agentic pipelines that search across heterogeneous sources, extract evidence, reason over that evidence, and synthesize long-form reports with citations \citep{venkit2025deeptrace}. These systems are increasingly used in enterprise workflows such as sales pitch generation, market and competitor analysis, and customer research \citep{choubey2026edr}. In these settings, the final report is the primary artifact used by downstream decision makers. A useful report must therefore do three things simultaneously: include the decision-relevant facts available in the collected evidence, attribute claims to the right sources, and remain maintainable when only a small part of the underlying evidence changes.

Existing DR benchmarks primarily score final reports with holistic rubrics~\citep{deepresearchbench2025,liveresearchbench2025,sharma2025researchrubricsbenchmarkpromptsrubrics}. While useful for overall quality, these scores often hide claim-level failures: reports may omit key facts, cite sources that do not support their claims, or introduce unsupported inferences. In enterprise DR, such errors are consequential, yet rubric scores provide little visibility into whether claims are supported, correctly attributed, and complete with respect to the available evidence \citep{ji2023hallucination,liu2023verifiability,narayananvenkit2024falsepromise}.

We address this gap with \probe{}, a claim-level auditing framework for DR reports. \probe{} decomposes each report into claims, checks them against the collected source evidence, and measures four dimensions: \textbf{hallucination}, \textbf{misattribution}, \textbf{citation hygiene}, and \textbf{necessary-fact recall}. It then aggregates these judgments into report-level metrics while preserving claim-level diagnostics, making it possible to identify which claims failed and why. The evaluator is an LLM judge calibrated through a three-track human study and grounded in explicit evidence comparisons \citep{zheng2023judging}.
% Using \probe{}, we find that strong DR pipelines exhibit systematic claim-level failures that rubric-based evaluations largely miss. Existing writers often omit necessary facts, cite sources that only partially support a claim, or introduce unsupported content despite producing fluent, well-organized reports. These errors suggest that improving DR reports requires a writer that explicitly tracks source-grounded claims, rather than relying on direct generation from retrieved evidence.
Using \probe{}, we find that strong DR pipelines still miss key facts, misattribute evidence, and introduce unsupported claims despite producing fluent reports. These failures point to a limitation of direct report generation from retrieved evidence: facts can lose structure and provenance during writing.

Motivated by this finding, we introduce \drwriter{}, a hierarchical claim-based writer that replaces only the report-generation module of existing DR systems. \drwriter{} first organizes source-grounded facts into an explicit claim hierarchy and then drafts the report from this structure. By separating evidence organization from report generation, \drwriter{} improves recall of important facts while preserving links between claims and their supporting sources.

We evaluate \drwriter{} by replacing only the original writer in three DR hosts: Enterprise Deep Research \citep{choubey2026edr}, NVIDIA \aiq{} \citep{nvidia_aiq}, and OpenDeepResearch \citep{langchain2025odr}. Upstream search and evidence collection are kept fixed. Across both evaluation settings, \drwriter{} consistently improves claim-level quality over the original writers, reducing hallucination and misattribution while increasing necessary-fact recall. These gains are largely hidden by rubrics-based scores, which move only slightly.

Beyond one-shot generation, \drwriter{} also supports low-cost report maintenance when source evidence changes. It uses the claim hierarchy to map changed source facts to affected claims, revise only the corresponding sections, and preserve the rest of the report. This matches enterprise settings, where the task and report structure often remain stable while individual facts evolve \citep{mukherjee2004enterprise}. Across different levels of source churn, our update system propagates changed facts into the report at the highest rate among update methods, while keeping output-token cost comparable to a single diff-based rewrite.

In summary, we make three main contributions. \textbf{(1)} We introduce \probe{}, a calibrated claim-level auditing framework for measuring faithfulness, attribution, citation quality, and coverage in DR reports. \textbf{(2)} We introduce \drwriter{}, a hierarchical claim-based writer that delivers superior claim-level report quality across DR hosts and writer models. \textbf{(3)} We develop an incremental revision method that uses claim structure to update reports efficiently and preserve unaffected content as source changes.

\section{\probe{}: A Claim-Level Report Auditing Framework}
\label{sec:probe}

% \probe{} is a \textit{four-phase pipeline} that takes a tool-log JSON and a markdown report and returns per-claim verdicts, resulting in four report-side measures.

% \subsection{Evaluated Properties}
% Following documented failure modes in commercial answer engines~\citep{narayananvenkit2024falsepromise,liu2023verifiability}, and using the atomic-decomposition view of long-form factuality probes~\citep{min2023factscore,wei2024longform}, we audit four properties that existing evaluations do not isolate:
% \begin{description}[leftmargin=1.4em,itemsep=1pt,topsep=2pt]
%     \item[Hallucination (\hall, $\downarrow$).] Does the report make claims that no retrieved source supports?
%     \item[Misattribution (\mis, $\downarrow$).] When the report cites source $k$, does source $k$ actually support the claim, or is the claim only supported by some other retrieved source?
%     \item[Uncited-but-supported (\unc, context).] Does the report omit citations on claims that retrieved sources do support? A high \unc{} together with low \hall{} and \mis{} indicates the writer is grounded but not always citing.
%     \item[Necessary fact recall (\rec, $\uparrow$).] Of the source facts that are highly relevant to the task, what fraction appear in the report?
% \end{description}

\probe{} is a \textit{four-phase pipeline} that takes the outputs collected through tool calls, such as web search, together with a report, evaluates each report claim against the collected evidence, and computes four report-level measures designed to capture failure modes identified in commercial answer engines~\citep{narayananvenkit2024falsepromise,liu2023verifiability} and prior work on long-form factuality evaluation~\citep{min2023factscore,wei2024longform}.
\begin{description}[leftmargin=1.4em,itemsep=1pt,topsep=2pt]
\item[Hallucination (\hall $\downarrow$).] Does the report make claims that no retrieved source supports?
\item[Misattribution (\mis $\downarrow$).] When the report cites source $k$, does source $k$ actually support the claim, or is the claim only supported by some other retrieved source?
\item[Uncited-but-supported (\unc).] Does the report omit citations for claims that retrieved sources do support? A high \unc{} together with low \hall{} and \mis{} indicates that the writer is grounded but does not always cite.
\item[Necessary fact recall (\rec $\uparrow$).] Of the source facts that are highly relevant to the task, what fraction appear in the report?
\end{description}

\begin{figure*}[t]
\centering
\resizebox{\linewidth}{!}{%
\begin{tikzpicture}[
  font=\scriptsize, every node/.style={font=\scriptsize},
  hostbox/.style ={draw=pdblue!70,  fill=pdblue!10,  line width=0.4pt, rounded corners=2pt,
                   text width=2.7cm, minimum height=7mm, align=center, inner sep=2pt},
  hostfade/.style={hostbox, draw=pdgray!45, fill=pdgray!6, text=pdgray!70},
  writerbox/.style={draw=pdgreen!75, fill=pdgreen!14, line width=0.4pt, rounded corners=2pt,
                    text width=2.3cm, minimum height=6.5mm, align=center, inner sep=2pt},
  sourcebox/.style={draw=pdgray!85, fill=pdgray!8, line width=0.4pt, rounded corners=4pt,
                    text width=2.6cm, minimum height=9mm, align=center, inner sep=1.5pt},
  factbox/.style ={draw=pdgold!85,  fill=pdgold!16,  line width=0.5pt, rounded corners=2pt,
                   text width=3.5cm, minimum height=9mm, align=center, inner sep=1.5pt},
  consbox/.style ={draw=pdpurple!75, fill=pdpurple!10, line width=0.4pt, rounded corners=2pt,
                   text width=3.0cm, minimum height=9mm, align=center, inner sep=2pt},
  evalbox/.style ={draw=pdorange!85, fill=pdorange!12, line width=0.4pt, rounded corners=2pt,
                   text width=3.0cm, minimum height=9mm, align=center, inner sep=2pt},
  grpfit/.style  ={draw=pdgray!45, dashed, rounded corners=3pt, inner sep=3.5pt},
  grouplbl/.style={font=\tiny\bfseries, color=pdgray},
  arr/.style     ={-{Latex[length=1.6mm]}, semithick, draw=pdgray!80},
  edgelbl/.style ={font=\tiny\itshape, color=pdgray, fill=white, inner sep=1pt}
]
\node[hostfade] (edr) {\textbf{\edrlong}};
\node[hostbox, below=2mm of edr] (aiq) {\textbf{NVIDIA \aiq}};
\node[hostfade, below=2mm of aiq] (odr) {\textbf{\odrlong}};
\node[grpfit, fit=(edr)(aiq)(odr)] (hostgrp) {};
\node[grouplbl, anchor=south west] at (hostgrp.north west) {Host deep-research pipelines (interchangeable)};

\node[sourcebox, right=8mm of hostgrp] (src)
  {\textbf{Source facts}\\\tiny self-contained, scored, deduplicated};
\node[factbox, right=8mm of src] (fg)
  {\textbf{Report fact graph}\\\tiny topics, sections, \texttt{\{\{CITE:fN\}\}} markers};
\node[consbox, right=9mm of fg] (rep)
  {\textbf{Assembled report}\\\tiny \texttt{report.md} with numbered citations $[N]$};

\node[evalbox, above=7mm of rep]              (bench)
  {\textbf{\probe{} audit}\\\tiny \hall \;\mis \;\unc \;\rec\\\tiny ($+$\,strict, $+$\,direct)};
\node[consbox, below=8mm of fg]               (upd)
  {\textbf{Fact-level updates}\\\tiny fact-patch, fact-diff\\\tiny (touch only changed facts)};
\node[consbox, right=11mm of upd]             (uprep)
  {\textbf{\textcolor{red}{Updated} report}\\\tiny \texttt{report.md} with numbered citations $[N]$};
\node[grpfit, draw=pdgreen!55, fit=(src)(fg)(rep)(upd)(uprep)] (writergrp) {};
\node[grouplbl, color=pdgreen!45!black, anchor=south west] at (writergrp.north west) {\drwriter{}};

\draw[arr] (hostgrp.east) -- node[edgelbl, pos=0.45, above=1mm] {sources} (src.west);
\draw[arr] (src.east) -- (fg.west);
\draw[arr] (fg.east) -- node[edgelbl, pos=0.55, sloped, above=1pt] {assemble} (rep.west);
\draw[arr] (bench.south |- writergrp.north) -- node[edgelbl, pos=0.5, right=1mm] {evaluate} (bench.south);
\draw[arr] (src.south) -- ++(0,-4mm) -| ([xshift=-3mm]upd.north);
\draw[arr] (fg.south) -- node[edgelbl, pos=0.35, right=3mm] {source delta} (upd.north);
\draw[arr] (rep.south) -- ++(0,-4mm) -| ([xshift=3mm]upd.north);
\draw[arr] (upd.east) -- (uprep.west);
\end{tikzpicture}%
}
\caption{\textbf{System overview.} A selected host deep-research pipeline supplies sources to \drwriter{}, which extracts source facts, materializes a report fact graph with \texttt{\{\{CITE:fN\}\}} markers, and assembles the user-facing \texttt{report.md}. \probe{} audits this fact-linked representation (Section~\ref{sec:probe}); when sources change, \drwriter{} applies fact-level updates and emits an updated report (Section~\ref{sec:exp:dyn}).}
\label{fig:pipeline}
\end{figure*}
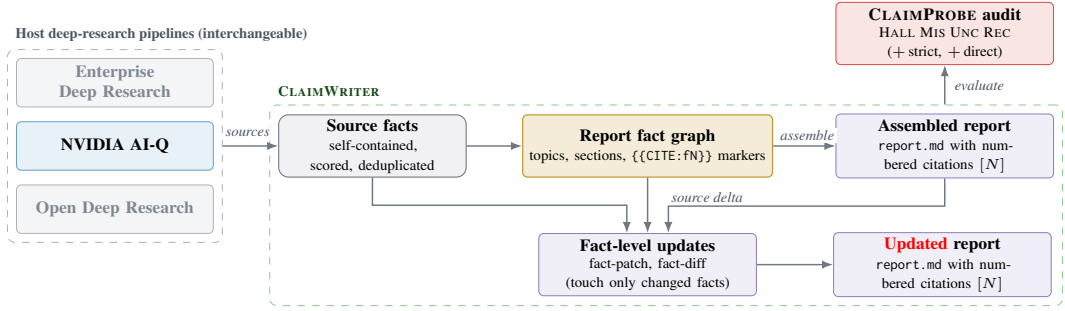

\subsection{Report Audit Procedure}
\probe{} %operates in four steps. It
first processes the collected source outputs and the report separately, then aligns report claims with source facts, and finally computes report-level measures.

% \textit{(1) Extract} parses tool logs into a canonical source table and segments the report twice: \texttt{report\_units} (sentence units, table summaries, contextualised table cells) for \rec{}, and \texttt{report\_claims} (atomic claims) for \hall, \mis, \unc{}.
% \textit{(2) Relevance} labels each source fact as \textsc{High}, \textsc{Related}, or \textsc{Unrelated} given the task, so the \rec{} denominator is the \textsc{High}-relevance set.
% \textit{(3) Alignment} embeds claims and facts with \texttt{text-embedding-3-small}, retrieves the top-$K{=}20$ candidates by cosine similarity, and asks a direction-aware judge for a verdict in \{\textsc{Supported}, \textsc{Partially Supported}, \textsc{Not Supported}\}. Report-to-source is strict (numeric precision required); source-to-report is lenient (paraphrase and subsumption count as coverage).
% \textit{(4) Metrics} computes the four rates from cached verdicts, plus strict variants where only fully \textsc{Supported} counts (\hall$_{\text{str}}$, \rec$_{\text{str}}$) and a direct-claim restriction \hall$_{\text{dir}}$. Details are in Appendix~\ref{app:probe}. The tasks are extracted from DeepResearchBench tasks published \cite{deepresearchbench2025}.

\textbf{Source processing:} converts collected source outputs into a standard format and extracts short, self-contained factual statements. It then labels each fact as \textsc{High}, \textsc{Related}, or \textsc{Unrelated} with respect to the task. We compute \rec{} using only \textsc{High}-relevance facts.

\textbf{Report processing:} parses the report's reference block to recover cited references. It then processes the report section by section, excluding the references section, and uses an LLM to extract report claim statements from each section together with their cited references.

\textbf{Alignment:} For each report claim, \probe{} embeds the claim and all source facts using \texttt{text-embedding-3-small}, retrieves the top-$K{=}20$ source facts by cosine similarity, and uses an LLM judge to label the claim as \textsc{Supported}, \textsc{Partially Supported}, or \textsc{Not Supported}. For recall, it applies the same retrieval-and-judging procedure in the reverse direction, starting from each source fact and matching it to report claims. Claim-to-source matching is strict and requires precise factual support, including for numeric values. Source-to-report matching is more lenient and treats paraphrase, summarization, and coverage as sufficient when the core information is preserved.

% \paragraph{Metrics:} Using the cached alignment judgments, \probe{} computes the four report-level measures, along with strict variants of hallucination and necessary-fact recall (\hall$_{\text{str}}$, \rec$_{\text{str}}$) in which only fully \textsc{Supported} matches count. Appendix~\ref{app:probe} provides further details on the metric definitions and human calibration.
\textbf{Metrics Computations:} Using the cached alignment judgments, \probe{} computes the four report-level measures. For report-to-source metrics, \hall{} treats a report claim as hallucinated only when no retrieved source fact supports it, i.e., when the claim is labeled \textsc{Not Supported}; \hall${\text{str}}$ is stricter and treats both \textsc{Partially Supported} and \textsc{Not Supported} report claims as not fully grounded. For source-to-report metrics, \rec{} credits a \textsc{High}-relevance source fact when the report fully or partially covers it, while \rec${\text{str}}$ credits only fully covered source facts. Thus, the default metrics count both \textsc{Supported} and \textsc{Partially Supported} as support, but in opposite alignment directions: report claims for \hall{} and source facts for \rec{}. Appendix~\ref{app:probe} provides further details on the metric definitions and human calibration.

\textbf{Judge calibration.}
We calibrate GPT-5.4-mini against human annotations. Three annotators labeled disjoint pools drawn from a stratified task sample (n=100), on hallucination, misattribution, and relevance, each in a blind setting. The AI judge agrees with the unanimous-human label (Cohen's $\kappa = 0.484, 0.797, 0.743$; details in Appendix~\ref{app:human}). The judge is asked to score one claim against a short shortlist at a time, the regime under which LLM-as-judge calibration is most stable \citep{zheng2023judging}.
% \pkc{move this to Appendix~\ref{app:probe}?}

\section{\drwriter{}: Hierarchical Claim-Based Writer Module}
\label{sec:writer}

% A claim-centric audit naturally suggests a claim-centric writer: rather than concatenating tool outputs and asking a long-context LLM to synthesise them in one pass, \drwriter{} maintains atomic source facts and atomic report claims as first-class objects, organises them into a topic structure before drafting, and emits prose with explicit \emph{claim-to-source linkage} that is resolved at assembly time \citep{gao2023enabling}. We study two ways of producing the topic structure.
A claim-centric audit naturally motivates a claim-centric writer. Instead of concatenating retrieved content and using a long-context LLM to synthesize a report, \drwriter{} maintains \emph{source facts} as first-class objects. It organizes these facts into a structured topic hierarchy before drafting and generates report text with explicit \emph{claim-to-source linkage}, which it resolves only in the final report construction stage \citep{gao2023enabling}.

% \paragraph{Top-down (\drtop{}).}
% A seven-stage pipeline splits into an idempotent preparation phase (stages 1--4) and a writing-and-review phase (stages 5--7) that retries only failing sections.
% \textit{(1)~Source loading} deduplicates tool logs into \texttt{pages.json}.
% \textit{(2)~Outline generation} drafts a flat list of topics from the user query \emph{without showing any fact evidence}, then critiques and revises. Producing the outline before evidence is deliberate: perturbing a source never forces an outline rebuild.
% \textit{(3)~Fact extraction} runs one LLM call per page in parallel and emits up to twenty-five self-sufficient facts per page (named subjects, exact numbers, explicit time and scope, no pronouns), scored on relevance and specificity. Facts are deduplicated and the top-$K$ retained with stable ids \texttt{f0000}\ldots\ in \texttt{facts.json}.
% \textit{(4)~Fact assignment} routes each fact to exactly one outline topic.
% \textit{(5)~Section writing} is a checkpointed per-section state machine that plans 2--5 beats and drafts prose with \texttt{\{\{CITE:fN\}\}} fact-id markers, not numbers.
% \textit{(6)~Assembly} concatenates sections, globally renumbers citations into a sequential $[N]$ space, and emits \texttt{report.md} plus an HTML rendering with a \texttt{\#\# References} list.
% \textit{(7)~Review} runs programmatic citation-integrity checks and an LLM coherence score per section to drive the retry loop.

Specifically, \drwriter{} first \textit{deduplicates} the retrieved source content and generates an initial topic outline directly from the user query, without conditioning on extracted evidence. This choice prevents partial retrieval coverage from determining the report structure and keeps the outline aligned with the information requirements of the query. It then \textit{extracts self-contained facts} from each source page in parallel, scores them for relevance and specificity, deduplicates them, and assigns each fact to a single outline topic. Given the resulting organization of topics and facts, \drwriter{} \textit{drafts each section independently} using fact-level citation markers (e.g., \texttt{{{CITE:fN}}}) that are converted into numeric references only when the report is constructed. The system then constructs the \textit{final report} by concatenating sections and resolving these into a global citation list.

\textbf{Efficient Handling of Source Changes}
% \label{sec:updates}

\drwriter{} produces a fact-linked report representation in which each extracted fact is tied to a topic assignment, a citation marker, and the section it supports. This representation allows the system to localize updates to only those sections affected by changes in source content, unlike current deep research pipelines that require a complete rerun of the process.

For each changed source, \drwriter{} compares the previous and updated source content together with the facts extracted from the previous version of that source. It then determines which existing fact ids should be updated or removed and which new facts should be added. For updated or removed facts, \drwriter{} uses the existing fact-to-topic assignments to identify the affected outline topics and revises the corresponding report sections. For newly added facts, \drwriter{} assigns each fact to the appropriate topic in the existing outline and updates the corresponding section to incorporate the new information.

This update procedure preserves the report structure while restricting rewriting to sections whose assigned facts changed, reducing token cost and avoiding unnecessary edits to unaffected text.

\begin{table}[t]
\centering
\small
\setlength{\tabcolsep}{3pt}
\renewcommand{\arraystretch}{1.18}
\begin{tabular}{@{}lccccc@{}}
\toprule
Host $+$ writer & comp. & insi. & inst. & read. & \textbf{ov.} \\
\midrule
\edr{}                     & .556 & .548 & .543 & .510 & .544 \\
\edr{} $+$ \drwriter{}     & .553 & .545 & .537 & .470 & .536 \\
\edr{} $+$ \drbot{}        & .554 & .550 & .539 & .466 & .538 \\
\midrule
\aiq{}                     & .558 & .557 & .554 & .528 & .553 \\
\aiq{} $+$ \drwriter{}     & .550 & .544 & .536 & .473 & .535 \\
\aiq{} $+$ \drbot{}        & .556 & .552 & .541 & .469 & .540 \\
\midrule
\odr{}                     & .546 & .539 & .544 & .523 & .540 \\
\odr{} $+$ \drwriter{}     & .543 & .534 & .534 & .485 & .530 \\
\odr{} $+$ \drbot{}        & .546 & .542 & .536 & .472 & .533 \\
\bottomrule
\end{tabular}
\caption{RACE scores ($n{=}100$, judge GPT-5.5). Columns are comprehensiveness, insight, instruction-following, readability, and overall (higher is better).
% Deltas under RACE are within $\pm 0.01$ on \edr{} and \aiq{}, the same task suite on which \probe{} (Table~\ref{tab:probe}) records up to $4\times$ swings in faithfulness and recall.
}
\label{tab:race}
\end{table}

\section{Experiments}
\label{sec:exp}
We report three main experiments that address the key questions an industry adopter would ask before deploying a deep-research stack: (i) Can the writer preserve holistic report quality under an established rubric (Section~\ref{sec:exp:race})? (ii) Can the writer retain necessary facts without fabricating new ones, across host pipelines and writer models (Section~\ref{sec:exp:probe})? and (iii) Can the system update reports quickly, consistently, and accurately when sources change (Section~\ref{sec:exp:dyn})?

\textbf{Setup.}
We conduct our experiments on DeepResearch Bench~\citep{deepresearchbench2025}, which contains 50 English and 50 Chinese deep research tasks; the task scope for each evaluation is specified below. We use three open-source deep research pipelines as host systems for source consolidation: Enterprise Deep Research (\edr{})~\citep{choubey2026edr}, NVIDIA \aiq{}~\citep{nvidia_aiq}, and OpenDeepResearch (\odr{})~\citep{langchain2025odr}. For each pipeline, we replace its original report-writing step with the proposed \drwriter{} to generate the final report. We use GPT-5.5 as the underlying model for all pipeline components, including report generation.

\textbf{Evaluation.}
For \probe{}-based evaluation, we use \texttt{text-embedding-3-small} as the embedding model, set retrieval to $K{=}20$ in both directions, and use a single LLM judge. We evaluate \probe{} with two backbone LLMs: GPT-5.4-mini on all 100 tasks (ids 1--100) and GPT-5.5 on a 10-task subset (ids 1--5, 51--55). All results in Table~\ref{tab:probe} use microaveraging, computed by pooling numerators and denominators across a run's tasks. We also report RACE scores~\citep{deepresearchbench2025} on all 50 English tasks using the original GPT-5.5 judge configuration.

% For the dynamic-update experiments, we use five tasks (ids 51--55) and evaluate two source-change budgets: 5 changed sources and 25 changed sources. In each setting, we update one specific fact per changed source document. As shown in Table~3, we report wall-clock update time, output-token cost, and update yield. We define yield as update coverage multiplied by update accuracy: the percentage of source changes reflected in the report times the accuracy of the updated report content. Since deep-research report updates are typically scheduled jobs rather than latency-critical interactive calls, we treat wall-clock time as a secondary measure and optimize primarily for output-token cost.

%\hh{mention model for incremental experiment over there.}%Dynamic update results in Table~\ref{tab:dyn} use GPT-5.4 over deepresearchbench tasks. All numbers in Table~\ref{tab:probe} are micro-averages (pooled numerator over pooled denominator across a run's tasks); rates in other tables are means.

\textbf{Baseline systems.}
We compare \drwriter{} against the system-default report writer step (\textbf{Baseline}). For each DR pipeline, we run the full pipeline to obtain the final report.
We also instantiate a naive variant of \drwriter{}, namely \textbf{\drbot{}}, that follows a bottom-up writing strategy:
it performs agglomerative clustering on the source facts to form topic groups, prunes query-irrelevant groups, then uses the remaining groups and the query to derive the report outline.
Finally, relevant facts to the outline are assigned and the report is generated for each section.
% Like \drwriter{},  this baseline represents the resulting report as a fact graph with fact-level citation markers (e.g., \texttt{{{CITE:fN}}}), which are resolved into final numeric references when the report is constructed.

% For the dynamic-update experiments in Table~\ref{tab}, we compare against two update baselines. The first is \textbf{\aiq{} native (replay)}, which stores the full DR trace and replays the pipeline after replacing only the changed source contents with their updated versions. The second is \textbf{Base (diff)}, which compares each old source document with its updated version, extracts the changed facts, aggregates all extracted changes, and updates the existing report in a single diff-based rewrite.

\section{Results}
\subsection{RACE scores stay close under \drwriter{}}\label{sec:exp:race}
We show the RACE scores for the baseline and \drwriter{} in Table~\ref{tab:race}.
On this report-level rubric, swapping in the claim-based writer leaves comprehensiveness and insight essentially unchanged but lowers readability (by $3.8$--$5.9$ points), yielding a small net drop in Overall ($\leq 1.8$ points) in every host.
This is a positive result: the claim-based writer largely preserves the holistic qualities that RACE measures while delivering the claim-level gains in Table~\ref{tab:probe}.

\subsection{\probe~reveals claim-level report quality differences}
\label{sec:exp:probe}
We report the \probe{} results for the $3{\times}3{\times}2$ (Host, Writer, Judge model) combinations in Table~\ref{tab:probe}.
Unlike RACE, \probe{} reveals large discrepancy between writers: the same writer swap that moves holistic scores only slightly can sharply reduce hallucination and misattribution while increasing necessary-fact recall.
For example, \drwriter{} reduces \hall{} from $15.89$ to $5.02$ and \mis{} from $18.94$ to $5.43$, while improving \rec{} from $36.83$ to $45.85$ on \edr{} with GPT-5.4-mini.
Across hosts and judge models, \drwriter{} and \drbot{} are consistently the strongest writers, trading the lead across cells, while the baseline is weakest.
We observe that GPT-5.4-mini is sufficient for our \probe{} setup: it gives nearly the same relative ranking of report writers as GPT-5.5.
However, absolute scores requires cautious interpretation since they differ moderately between the judge models.
Since all writers share a single backbone model (GPT-5.5), any judge self-preference due to matching generator \& evaluator pairs applies to all cases equally.
We verify this by also running \probe{} with Gemini-3.5-flash as the backbone and demonstrate the same relative system rankings in Appendix~\ref{app:gemini}.
% The uncited-but-supported rate (\unc{}) is contextual: it often rises under \drbot{} or \drwriter{}, but this accompanies lower \hall{}/\mis{} and higher \rec{}, indicating a residual citation-hygiene gap rather than a standalone factuality failure.
\begin{table*}[t]
\centering
\scriptsize
\setlength{\tabcolsep}{3pt}
\renewcommand{\arraystretch}{1.18}
\begin{tabular}{@{}ll|cccccc|cccccc@{}}
\toprule
& & \multicolumn{6}{c|}{\textbf{GPT-5.5} ($n{=}10$)} & \multicolumn{6}{c}{\textbf{GPT-5.4-mini} ($n{=}100$)} \\
Host & Writer & \hall{}$\downarrow$ & \hall$_{\text{str}}$$\downarrow$ & \mis{}$\downarrow$ & \rec{}$\uparrow$ & \rec$_{\text{str}}$$\uparrow$ & \unc{} & \hall{}$\downarrow$ & \hall$_{\text{str}}$$\downarrow$ & \mis{}$\downarrow$ & \rec{}$\uparrow$ & \rec$_{\text{str}}$$\uparrow$ & \unc{} \\
\midrule
\multirow{3}{*}{\aiq}
& Baseline     & \prLow{14.84} & \prLow{78.35} & \prLow{24.68} & \prLow{36.08} & \prLow{\phantom{0}7.85} & 83.23 & \prLow{23.27} & \prLow{73.51} & \prLow{22.31} & \prLow{51.35} & \prLow{20.59} & 73.83 \\
& $+$\drbot    & \prMid{\phantom{0}3.41} & \prHigh{36.80} & \prMid{\phantom{0}5.29} & \prMid{56.31} & \prMid{28.46} & 88.92 & \prHigh{\phantom{0}5.08} & \prHigh{28.09} & \prMid{\phantom{0}4.13} & \prHigh{68.71} & \prMid{46.59} & 79.61 \\
& $+$\drwriter{} & \prHigh{\phantom{0}3.37} & \prMid{39.02} & \prHigh{\phantom{0}5.10} & \prHigh{56.99} & \prHigh{29.97} & 88.46 & \prMid{\phantom{0}5.20} & \prMid{28.49} & \prHigh{\phantom{0}3.62} & \prMid{68.29} & \prHigh{47.75} & 80.61 \\
\midrule
\multirow{3}{*}{\odr}
& Baseline     & \prLow{23.80} & \prLow{75.67} & \prLow{\phantom{0}7.67} & \prLow{33.61} & \prLow{\phantom{0}9.21} & 66.68 & \prLow{27.12} & \prLow{67.08} & \prLow{\phantom{0}5.92} & \prLow{59.34} & \prLow{32.82} & 63.60 \\
& $+$\drbot    & \prHigh{\phantom{0}5.16} & \prHigh{42.34} & \prHigh{\phantom{0}2.23} & \prHigh{63.79} & \prHigh{43.89} & 78.06 & \prHigh{\phantom{0}6.78} & \prHigh{31.16} & \prHigh{\phantom{0}2.28} & \prHigh{80.48} & \prHigh{67.79} & 73.38 \\
& $+$\drwriter{} & \prMid{\phantom{0}5.92} & \prMid{48.90} & \prMid{\phantom{0}2.38} & \prMid{57.82} & \prMid{36.58} & 78.04 & \prMid{\phantom{0}8.78} & \prMid{35.87} & \prMid{\phantom{0}2.50} & \prMid{75.93} & \prMid{61.13} & 72.60 \\
\midrule
\multirow{3}{*}{\edr}
& Baseline     & \prLow{12.09} & \prLow{61.86} & \prLow{24.23} & \prLow{21.12} & \prLow{\phantom{0}3.92} & 85.11 & \prLow{15.89} & \prLow{54.05} & \prLow{18.94} & \prLow{36.83} & \prLow{13.67} & 77.75 \\
& $+$\drbot    & \prHigh{\phantom{0}3.91} & \prHigh{37.67} & \prMid{\phantom{0}6.89} & \prHigh{31.72} & \prMid{11.73} & 91.14 & \prHigh{\phantom{0}4.91} & \prMid{28.56} & \prMid{\phantom{0}5.88} & \prMid{43.99} & \prMid{22.69} & 82.71 \\
& $+$\drwriter{} & \prMid{\phantom{0}4.65} & \prMid{39.31} & \prHigh{\phantom{0}6.77} & \prMid{31.56} & \prHigh{11.79} & 86.38 & \prMid{\phantom{0}5.02} & \prHigh{28.55} & \prHigh{\phantom{0}5.43} & \prHigh{45.85} & \prHigh{24.48} & 83.57 \\
\bottomrule
\end{tabular}
\caption{\textbf{\probe{} audit on the host $\times$ writer $\times$ judge-model factorial.} All values are percentages, micro-averaged over pooled (numerator, denominator) pairs. Green cells are the best writer in each (host, judge-model) block; \unc{} is contextual and left uncolored.}
\label{tab:probe}
\end{table*}

\section{Discussion}
\label{sec:discussion}

% \paragraph{Why reports matter.} Three findings carry the industry argument.

% First, \emph{writing structure substitutes for raw model capability}. With the small writer model, $+$\drtop{} reaches \hall{} of $5.87$--$11.16$ and \rec{} of $59.28$--$64.32$ across the three hosts, matching or exceeding the strong writer model under the host baselines (\hall{} $6.93$--$17.44$, \rec{} $43.92$--$46.61$). \textbf{Changing the writer, not the model, closes most of the gap}, and the small model is an order of magnitude cheaper at inference.

% Second, \textit{the gains generalise across hosts and writer models}. $+$\drtop{} wins every metric on every host for \emph{both} models, with tight cross-host spreads on GPT-5.5 (\hall{} $2.5$--$11.7$, \mis{} $8.1$--$10.2$, \rec{} $59.1$--$61.1$); the improvement is not an artifact of any one retrieval substrate. Strict-recall (\rec$_{\text{str}}$) confirms this in a tougher form: $+$\drtop{} roughly triples the fraction of high-relevance facts that pass the strict-citation test.

% Third, \textit{\drbot{} is the lower-variance fallback}, never the leader. Its cross-host \hall{} spread is the tightest of the three writers ($1.2$--$1.6$ pp), and it roughly halves baseline misattribution on every cell, but it never overtakes $+$\drtop{} on coverage or precision. We treat \drtop{} as the default and report \drbot{} where the contrast is informative.

\subsection{Holistic quality and source faithfulness can diverge}

The RACE result in Table~\ref{tab:race} and the \probe{} result in Table~\ref{tab:probe} show that surface-level report quality and source faithfulness are partly separable.
A writer can keep the holistic evaluation score close to the baseline while changing the claim-to-source relationship substantially.
In other words, the broad report-quality measures are useful but insufficient in practice: they should be paired with a claim-level audit that checks whether the report cites the right sources and retains the necessary facts.

\textbf{Why claim-level judging is better scoped.}
\probe{} also uses an LLM judge, but each decision is local. A \hall{} or \mis{} verdict compares one report claim against a top-$K{=}20$ evidence shortlist; a \rec{} verdict asks whether one \textsc{High}-relevance source fact is paraphrased or subsumed in the report. Since the judge never evaluates either artifact end to end, page-level confounds such as long-prose anchoring, length bias, and presentation effects are reduced. This matches the observation from \citet{zheng2023judging} that LLM-as-a-judge is more reliable under tightly scoped prompts, and our human study (Section~\ref{sec:probe}) confirms higher agreement in this restricted setting.

\begin{table*}[t]
\centering
\scriptsize
\setlength{\tabcolsep}{4pt}
\renewcommand{\arraystretch}{1.18}
\begin{tabular}{@{}ll|rr|rr@{}}
\toprule
& & \multicolumn{2}{c|}{\textbf{5} sources perturbed} & \multicolumn{2}{c}{\textbf{25} sources perturbed} \\
Host & Edit strategy & Out Tok & \textsc{Yield} [\%] & Out Tok & \textsc{Yield} [\%] \\
\midrule
\aiq{} & native (replay)          & \prFloor{159K} & \prFloor{\phantom{0}0.0}  & \prFloor{354K} & \prFloor{\phantom{0}0.5} \\
\aiq{} & $+$Base (diff)           & \prHigh{\phantom{0}20K}   & \prLow{16.7}              & \prHigh{\phantom{0}21K}   & \prLow{\phantom{0}6.9}   \\
\edr{} & $+$Base (diff)           & \prMid{\phantom{0}68K}    & \prMid{40.5}              & \prMid{\phantom{0}70K}    & \prMid{30.9}             \\
\edr{} & $\mathbf{+}$\drwriter{}  & \prHigh{\phantom{0}34K}   & \prHigh{47.6}             & \prMid{\phantom{0}72K}    & \prMid{39.7}             \\
\edr{} & $\mathbf{+}$\drbot{}     & \prLow{\phantom{0}78K}    & \prMid{45.2}              & \prLow{100K}   & \prHigh{41.2}            \\
\bottomrule
\end{tabular}
\caption{\textbf{Dynamic-update cost and update yield.} Output tokens are per-report averages over the tasks. \textsc{Update Yield} is the \emph{clean} rate: the percentage of relevant perturbed facts whose new value reaches the regenerated report while the old value is removed (higher is better). $\mathbf{+}$\drwriter{} is the fact-diff (top-down) updater and $\mathbf{+}$\drbot{} the bottom-up fact-diff updater. Cell colour: green favourable, yellow mid, red unfavourable, dark-red catastrophic (the cost ceiling).}
\label{tab:dyn}
\end{table*}

\subsection{Dynamic Report Update: Yield and Cost}
\label{sec:exp:dyn}

Enterprise reports often keep the task and report structure fixed while the underlying sources drift. To model this setting, we use five DeepResearch Bench tasks (ids 51--55) and evaluate two source-change budgets: 5 changed sources and 25 changed sources. In each setting, we update one specific numerical fact per changed source document with GPT-5.5, covering years, quantities, trends, or combinations thereof, and validate each edit for consistency with the surrounding source context. Each perturbation defines a new (perturbed) and an old (original) target value; over the query-relevant facts, an LLM judge labels the regenerated report as containing the \textsc{New} value, the \textsc{Old} value, \textsc{Both}, or \textsc{Absent}.

We compare \drwriter{} against two update baselines: \textbf{\aiq{} native (replay)}, which stores the full DR trace and replays the pipeline after replacing only the changed source contents, and \textbf{Base (diff)}, which compares each old source document with its updated version, extracts the changed facts, aggregates them, and rewrites the existing report in a single diff-based update.

As shown in Table~\ref{tab:dyn}, we report output-token cost and update yield. We define yield as the \emph{clean} rate: the percentage of relevant perturbed facts for which the regenerated report states the new value \emph{and} no longer states the old one. Since deep-research report updates are typically scheduled jobs rather than latency-critical interactive calls, we optimize primarily for output-token cost.

We find that update quality depends less on rewriting strength than on whether the system exposes a fine-grained edit handle. Full replay reruns the pipeline and sets the cost ceiling, but it still misses many small source changes and produces very low yield. A generic diff over a monolithic report reduces cost, but it lacks a stable object corresponding to the changed fact and often edits around the change rather than propagating it. In contrast, \drwriter{} casts source changes into fact changes and uses the constructed fact graph to localize revisions to affected sections. As a result, \drwriter{} achieves substantially higher update yield while staying close to diff-style output-token costs and preserving report faithfulness.\footnote{Refer to Appendix~\ref{app:dyn} for \probe{} evaluation on reports from perturbed source.}

\textbf{Steady-state cost note.}
\drwriter{} uses more first-run input tokens than the host's native writer because it re-extracts facts from the retrieved sources (Table~\ref{tab:writingcost} in Appendix~\ref{app:cost}). This one-time cost materializes the fact graph; subsequent updates amortize it through the lower incremental-update budget in Table~\ref{tab:dyn}. For workflows that update the same report multiple times, this amortization dominates. For one-shot reports, users should weigh the higher first-run cost against the benefits of claim-level structure and maintainability.

\subsection{Interpreting secondary metrics}

% \paragraph{A modest coverage--coherence trade-off.}
% Across all three hosts on GPT-5.5, the contradicting-claim rate rises monotonically from baseline through \drbot{} to \drtop{} (\aiq{}: $0.98$, $2.03$, $2.62$; \odr{}: $0.85$, $1.89$, $2.20$; \edr{}: $1.80$, $1.89$, $2.65$). Denser, higher-coverage reports surface more internal tension, but the absolute level stays below $3\%$ everywhere on the strong writer model. We read this as the expected cost of pulling more decision-relevant material into the report, not a regression, since fabrication (\hall{}) and misattribution (\mis{}) drop simultaneously.
% \hh{Omitting as we don't report the INC}

\textbf{Why high \unc{} is a benign signal.}
The uncited-but-supported rate (\unc{}) sits at $64$--$91\%$ everywhere and tends to \emph{rise} with \drwriter.
Read together with the simultaneous drops in \hall{} and \mis{} and the lifts in strict-recall (\rec$_{\text{str}}$), this means the uncited residue is increasingly composed of grounded background and connective tissue (general knowledge, framing sentences), not fabrications.
\unc{} alone should therefore be read as a citation-hygiene gap, not as a fabrication signal.

\section{Related Work}
\label{sec:related}

Production deep-research systems include Google Gemini Deep Research~\citep{google2025geminideepresearch}, OpenAI Deep Research~\citep{openai2025deepresearch}, and open-source efforts such as Enterprise Deep Research~\citep{choubey2026edr}, NVIDIA AI-Q~\citep{nvidia_aiq}, and Open Deep Research~\citep{langchain2025odr}. Existing systems emphasize planning, search, and stopping criteria, while treating report generation largely as a final long-context synthesis step. Recent benchmarks, including DeepResearch Bench~\citep{deepresearchbench2025}, LiveResearch Bench~\citep{liveresearchbench2025}, ResearchRubrics~\citep{sharma2025researchrubricsbenchmarkpromptsrubrics}, DeepResearch Bench II~\citep{li2026deepresearchbenchiidiagnosing}, DRBench~\citep{abaskohi2026drbenchrealisticbenchmarkenterprise}, and GAIA~\citep{mialon2024gaia}, evaluate final reports, full agents, or trajectories using rubrics, human-authored criteria, or ground-truth insight recall. These benchmarks improve coverage, realism, and enterprise grounding, but do not isolate the report writer under fixed retrieved evidence; as Section~5.1 shows, aggregate scores can mask large claim-level faithfulness differences.

A parallel line of work evaluates factuality and attribution below the document level. FActScore~\citep{min2023factscore} and SAFE~\citep{wei2024longform} decompose long-form generations into atomic facts or claims, while ALCE~\citep{gao2023enabling}, attributed QA~\citep{bohnet2022attributed}, and RAGAS~\citep{es2024ragas} evaluate citation quality or retrieval-grounded faithfulness. Prior studies also document hallucinations and citation errors in deployed search and QA systems~\citep{liu2023verifiability,narayananvenkit2024falsepromise,venkit2025deeptrace}, and surveys summarize hallucination phenomena more broadly~\citep{ji2023hallucination}. \probe{} extends this line to multi-source, citation-heavy deep-research reports, isolating writer-attributable failures through claim-level verdicts over fixed retrieved evidence.

\section{Conclusion}

We introduce \probe{}, a claim-level audit for DR reports, and \drwriter{}, a hierarchical claim-based DR report writer. Across three DR hosts, replacing only their writer with \drwriter{} preserves rubrics-based performance while improving claim-level faithfulness and recall. The same fact-linked representation supports localized report revision under source drift, producing more accurate updates at lower output-token cost. These results motivate focused work on DR report writing and update systems that preserve evidence structure, attribution, and maintainability. We have made public our pipeline and benchmark\footnote{https://github.com/SalesforceAIResearch/claimwriter-deep-research}.
% Together, these results suggest that deep-research systems should be evaluated at the report level and built around explicit claim graphs that support both fine-grained auditing and efficient updates.

\clearpage
\section*{Limitations}
Our evaluation has two main limitations. First, \probe{} uses an LLM judge. We mitigate this by scoring local decisions: one claim against a top-$K{=}20$ evidence shortlist, and by validating the judge against human annotations. Still, users who deploy \probe{} in new domains should recalibrate the judge on their own task distributions.

Second, our dynamic-update study uses a controlled source-drift setting: five tasks, two source-change budgets, and one numerical fact update per changed source. This design lets us isolate whether a system propagates changed facts into the report, but it is only a first step toward studying report maintenance under evolving evidence. We hope it motivates further research on broader update types, including new sources, removed evidence, qualitative changes, and updates that require restructuring the report. Finally, \drwriter{} pays a higher first-run cost to materialize the fact graph, so its cost advantage is strongest in workflows where reports are updated over time.

% \paragraph{Judge dependence, scoped tightly.}
% \probe{} uses a single LLM judge, the standard objection levelled at any LLM-based audit \citep{zheng2023judging}. We mitigate by chunking: each verdict is one claim against a top-$K{=}20$ embedding shortlist, never an entire report. Table~\ref{tab:human} shows that on this scoped task the AI judge agrees with unanimous-human verdicts better than humans agree with each other. The strict and direct variants further reduce judge laxity, and a \texttt{netloc + first two path segments} site-prefix fallback for \mis{} resolves common mega-domain false positives, at the cost of possibly under-flagging genuine misattribution between sibling pages of the same dataset. Downstream users should re-run the human-validation slice on their own task distribution.

% \paragraph{Inconsistency is recorded but not headlined.}
% Our harness also records an intra-report inconsistency count (pairwise judge over report claims), low for $+$\drtop{} on five of six head-to-head cells ($< 3\%$ on every host with the strong writer and on \aiq{}/\edr{} with the small writer). The one exception is GPT-5.4-mini on \odr{}, where every writer (including the host baseline) sits at $9$--$13\%$, marking it as a corpus--model interaction that warrants a consistency-repair pass before deployment. Inconsistency scoring semantics deserve a separate calibration study before promotion to a primary metric.

%\section*{Ethical Considerations}

\bibliographystyle{iclr2026_conference}
\bibliography{custom}

\clearpage

\appendix

\section{\probe{} pipeline details}
\label{app:probe}

\paragraph{Extract.} Tool logs become a normalized \texttt{sources} table with URL canonicalization (scheme, \texttt{www} prefix, trailing slash, and fragments removed) and a boilerplate stripper. The report's reference list is parsed by three regular expressions (canonical \texttt{[N] url}, numbered Markdown \texttt{N. [Title](url)}, and descriptive \texttt{[N] description: url}) to tolerate real-world citation conventions. The report is segmented twice: \texttt{report\_units.json} (sentence units, table summaries, contextualized table-cell atoms) for \rec{}, and \texttt{report\_claims.json} (claims) for \hall, \mis, \unc{}. Source contents are segmented into source facts by an LLM extractor instructed to make facts self-sufficient, replace pronouns with entity names, and drop boilerplate.

\paragraph{Relevance.} Each source fact receives a label in \{\textsc{High}, \textsc{Related}, \textsc{Unrelated}\} conditioned on the task statement. \textsc{High} facts directly answer the task or provide a central statistic the report is expected to use; \textsc{Related} facts provide background; \textsc{Unrelated} facts are off-topic. \textsc{Unrelated} facts are excluded from the embedding index.

\paragraph{Alignment.} For each query item we retrieve the top-$K{=}20$ candidates and ask a holistic judge for a strict JSON verdict in \{\textsc{S}, \textsc{PS}, \textsc{NS}\} with a 1-to-5 confidence score and a 20-word justification. The judge is direction-aware: the report-to-source prompt is strict (numeric precision required); the source-to-report prompt is lenient (paraphrase and subsumption count as coverage).

\paragraph{Metrics.} With $\mathcal{C}$ the set of report claims and $\mathcal{F}_{\text{high}}$ the \textsc{High} source-fact set,
{\small\begin{align*}
\hall      &= |\{c : \texttt{r2s}(c) {=} \textsc{NS}\}|\,/\,|\mathcal{C}| \\
\hall_{\text{str}} &= |\{c : \texttt{r2s}(c) {\neq} \textsc{S}\}|\,/\,|\mathcal{C}| \\
\mis       &= |\{c {\in} \mathcal{C}_{\text{cited}} : \neg\text{by\_cited}(c) \wedge \text{by\_other}(c)\}|\,/\,|\mathcal{C}_{\text{cited}}| \\
\unc       &= |\{c {\in} \mathcal{C}_{\text{uncited}} : \text{supported}(c)\}|\,/\,|\mathcal{C}_{\text{uncited}}| \\
\rec       &= |\{f {\in} \mathcal{F}_{\text{high}} : \texttt{s2r}(f) {\in} \{\textsc{S},\textsc{PS}\}\}|\,/\,|\mathcal{F}_{\text{high}}|.
\end{align*}}%
A site-prefix fallback (\texttt{netloc + first two path segments}) avoids inflating \mis{} on government and statistical mega-domains.

\section{Human validation of the LLM judge}
\label{app:human}

Three annotators labeled disjoint pools drawn from a stratified task sample. We report Cohen's $\kappa$ for human-human agreement and for the AI judge against the human majority and the unanimous-human labels.

\begin{table}[h]
\centering
\small
\setlength{\tabcolsep}{3pt}
\begin{tabular}{@{}llccc@{}}
\toprule
Study & Task & Hum.\,$\kappa$ & AI/maj & AI/unan. \\
\midrule
A & Hallucination   & $+0.173$ & $+0.299$ & $+0.484$ \\
B & Misattribution  & $+0.305$ & $+0.570$ & $+0.797$ \\
C & Relevance       & $+0.500$ & $+0.600$ & $+0.743$ \\
\bottomrule
\end{tabular}
\caption{Human and LLM judge agreement on the three judging tasks (judge GPT-5.4-mini).}
\label{tab:human}
\end{table}

\section{Robustness to the judge model}
\label{app:gemini}

The \probe{} judge we use was GPT-5.5, drawn from the same family as the backbone that writes the reports.
While this may raise a concern regarding self-preference in LLM judges, we note that such effect is applied to \textit{all} of them equally and leaves their relative ranking unchanged.
To show this, we also ran \probe{} with Gemini-3.5-flash as an independent judge on the 10-task subset (ids 1--5, 51--55), holding the writers, sources, and extraction steps fixed.
Table~\ref{tab:probe_gemini} demonstrates a similar observation: our proposed report writing systems significantly improve over the baselines in every metric.
On \aiq{}, for instance, replacing the baseline with \drbot{} drops \hall{} from $57.14$ to $19.81$ and \mis{} from $15.89$ to $0.86$, and raises \rec{} from $15.97$ to $41.22$.

% ClaimProbe audit, Gemini-3.5-flash judge (n=10; ids 1-5, 51-55).
% Mirrors tab:probe structure/coloring but for a single judge model.
% Micro-averaged over pooled (numerator, denominator). Green = best writer per
% (host) block, yellow = mid, red = worst; Unc is contextual and uncolored.
\begin{table}[h]
\centering
\scriptsize
\setlength{\tabcolsep}{3pt}
\renewcommand{\arraystretch}{1.18}
\begin{tabular}{@{}ll|cccccc@{}}
\toprule
& & \multicolumn{6}{c}{\textbf{Gemini-3.5-flash} ($n{=}10$)} \\
Host & Writer & \hall{}$\downarrow$ & \hall$_{\text{str}}$$\downarrow$ & \mis{}$\downarrow$ & \rec{}$\uparrow$ & \rec$_{\text{str}}$$\uparrow$ & \unc{} \\
\midrule
\multirow{3}{*}{\aiq}
& Baseline       & \prLow{57.14} & \prLow{83.69} & \prLow{15.89} & \prLow{15.97} & \prLow{\phantom{0}5.43} & 28.72 \\
& $+$\drbot      & \prHigh{19.81} & \prHigh{49.43} & \prHigh{\phantom{0}0.86} & \prHigh{41.22} & \prHigh{27.14} & 19.05 \\
& $+$\drwriter{} & \prMid{23.03} & \prMid{54.92} & \prMid{\phantom{0}0.99} & \prMid{39.71} & \prMid{25.89} & 22.86 \\
\midrule
\multirow{3}{*}{\odr}
& Baseline       & \prLow{58.72} & \prLow{81.94} & \prLow{\phantom{0}2.51} & \prLow{22.49} & \prLow{\phantom{0}9.25} & 23.25 \\
& $+$\drbot      & \prHigh{23.50} & \prHigh{54.96} & \prHigh{\phantom{0}0.27} & \prHigh{59.64} & \prHigh{46.72} & 35.71 \\
& $+$\drwriter{} & \prMid{29.08} & \prMid{58.45} & \prMid{\phantom{0}0.30} & \prMid{52.35} & \prMid{40.42} & 25.21 \\
\midrule
\multirow{3}{*}{\edr}
& Baseline       & \prLow{45.24} & \prLow{71.19} & \prLow{12.86} & \prLow{\phantom{0}9.65} & \prLow{\phantom{0}3.32} & 26.20 \\
& $+$\drbot      & \prHigh{20.57} & \prHigh{48.21} & \prHigh{\phantom{0}1.15} & \prHigh{17.23} & \prHigh{\phantom{0}9.91} & 26.67 \\
& $+$\drwriter{} & \prMid{25.27} & \prMid{55.51} & \prMid{\phantom{0}1.39} & \prMid{17.10} & \prMid{\phantom{0}9.80} & 21.74 \\
\bottomrule
\end{tabular}
\caption{\textbf{\probe{} audit with the Gemini-3.5-flash judge} ($n{=}10$; ids 1--5, 51--55). All values are percentages, micro-averaged over pooled (numerator, denominator) pairs. Green cells are the best writer in each host block; \unc{} is contextual and left uncolored.}
\label{tab:probe_gemini}
\end{table}

\section{Probe under perturbation}
\label{app:dyn}

We re-run \probe{} on every perturbed regeneration of Section~\ref{sec:exp:dyn}. Table~\ref{tab:incrprobe} shows that all \edr{} update methods (Base, FP, FD) cluster within $\pm 1$ percentage point of the unperturbed \edr{} baseline on every \probe{} metric; the surgical updates do not introduce drift. The \aiq{} systems hallucinate and misattribute markedly more in every configuration ($\hall_{\text{str}} \approx 66{-}68$, $\mis \approx 44{-}45$) and recall fewer \textsc{High} facts ($\rec \approx 31{-}38$), independently of the update strategy.

% \begin{table}[h]
% \centering
% \scriptsize
% \setlength{\tabcolsep}{3pt}
% \begin{tabular}{lcccccc}
% \toprule
% Method & \hall{} & \hall$_{\text{str}}$ & \mis{} & \rec{} & \rec$_{\text{str}}$ & \unc{} \\
% \midrule
% \edr{}              & $6.1$  & $43.7$ & $12.8$ & $54.6$ & $20.4$ & $99.0$ \\
% \edr{}-base-5       & $6.4$  & $44.0$ & $13.3$ & $54.6$ & $20.3$ & $98.2$ \\
% \edr{}-base-25      & $6.8$  & $43.9$ & $12.1$ & $54.6$ & $20.1$ & $98.9$ \\
% \edr{}-FP-5         & $6.4$  & $43.6$ & $12.7$ & $54.5$ & $20.2$ & $98.6$ \\
% \edr{}-FP-25        & $6.5$  & $45.2$ & $12.5$ & $53.9$ & $20.1$ & $97.1$ \\
% \edr{}-FD-5         & $6.2$  & $44.1$ & $12.6$ & $54.1$ & $20.0$ & $99.0$ \\
% \edr{}-FD-25        & $6.8$  & $45.3$ & $13.4$ & $54.4$ & $20.1$ & $96.9$ \\
% \aiq{}              & $10.4$ & $68.0$ & $44.5$ & $30.8$ & $8.4$ & $90.1$ \\
% \aiq{}-5  (native)  & $9.4$  & $67.1$ & $43.5$ & $37.9$ & $8.0$ & $89.5$ \\
% \aiq{}-25 (native)  & $10.1$ & $66.4$ & $44.6$ & $38.4$ & $8.3$ & $89.6$ \\
% \bottomrule
% \end{tabular}
% \caption{\probe{} on perturbed regenerations ($\%$, judge GPT-5.5). \edr{} update strategies cluster within $\pm 1$ of the unperturbed \edr{} baseline; updates do not drift.}
% \label{tab:incrprobe}
% \end{table}
\begin{table*}[h]
\centering
\scriptsize
\setlength{\tabcolsep}{3pt}
\begin{tabular}{llcccccc}
\toprule
\#Ptb. & Method & \hall{} & \hall$_{\text{str}}$ & \mis{} & \rec{} & \rec$_{\text{str}}$ & \unc{} \\
\midrule

\multirow{2}{*}{Baseline}
& \edr{}             & $6.1$  & $43.7$ & $12.8$ & $54.6$ & $20.4$ & $99.0$ \\
& \aiq{}             & $10.4$ & $68.0$ & $44.5$ & $30.8$ & $8.4$ & $90.1$ \\
\midrule

\multirow{4}{*}{Subset 5}
& \edr{}-base     & $6.4$  & $44.0$ & $13.3$ & $54.6$ & $20.3$ & $98.2$ \\
% & \edr{}-FP        & $6.4$  & $43.6$ & $12.7$ & $54.5$ & $20.2$ & $98.6$ \\
& \edr{}-\drwriter{}        & $6.2$  & $44.1$ & $12.6$ & $54.1$ & $20.0$ & $99.0$ \\
& \aiq{} (native)  & $9.4$  & $67.1$ & $43.5$ & $37.9$ & $8.0$ & $89.5$ \\
\midrule

\multirow{4}{*}{Subset 25}
& \edr{}-base     & $6.8$  & $43.9$ & $12.1$ & $54.6$ & $20.1$ & $98.9$ \\
% & \edr{}-FP       & $6.5$  & $45.2$ & $12.5$ & $53.9$ & $20.1$ & $97.1$ \\
& \edr{}-\drwriter{}       & $6.8$  & $45.3$ & $13.4$ & $54.4$ & $20.1$ & $96.9$ \\
& \aiq{} (native) & $10.1$ & $66.4$ & $44.6$ & $38.4$ & $8.3$ & $89.6$ \\
\bottomrule
\end{tabular}
\caption{\probe{} on perturbed regenerations ($\%$, judge GPT-5.5). \edr{} update strategies cluster within $\pm 1$ of the unperturbed \edr{} baseline; updates do not drift.}
\label{tab:incrprobe}
\end{table*}

\section{Steady-state writing cost}
\label{app:cost}

Per-report writing-stage tokens (model GPT-5.5), averaged over deepresearchbench tasks. Counts the LLM calls from processing retrieved tool output to the final report, including intermediate artifacts; excludes search and tool calls and query and plan generation.

\begin{table}[h]
\centering
\scriptsize
\setlength{\tabcolsep}{3pt}
\begin{tabular}{@{}lrr@{}}
\toprule
System          & in tok / report & out tok / report \\
\midrule
\edr{}                  & $1{,}019{,}276$   & $\phantom{00}92{,}252$ \\
\edr{} $+$\drwriter{}      & $9{,}697{,}825$   & $\phantom{0}700{,}626$ \\
%\edr{} $+$\drbot{}      & $15{,}475{,}435$ & $2{,}582{,}801$ \\
\aiq{}                  & $\phantom{0}882{,}649$ & $\phantom{0}344{,}646$ \\
\aiq{} $+$\drwriter{}      & $5{,}754{,}095$   & $\phantom{0}528{,}723$ \\
% \aiq{} $+$\drbot{}      & $8{,}909{,}963$   & $1{,}584{,}417$ \\
\odr{}                  & $2{,}331{,}138$   & $\phantom{0}293{,}372$ \\
\odr{} $+$\drwriter{}      & $2{,}779{,}813$   & $\phantom{0}248{,}320$ \\
% \odr{} $+$\drbot{}      & $4{,}089{,}806$   & $\phantom{0}707{,}591$ \\
\bottomrule
\end{tabular}
\caption{Steady-state writing-stage token cost (averages per report). $+$\drwriter{} and $+$\drbot{} re-extract facts from the entire retrieved corpus, so their input cost is higher than the host baselines (which only process what they synthesize). The fact-graph cost is paid once per first-run and amortizes against the incremental-update budget in Table~\ref{tab:dyn}.}
\label{tab:writingcost}
\end{table}

\section{Judge and extractor prompts}
\label{app:prompts}

We reproduce the system prompts used by the \probe{} judge and the extractors, taken verbatim from the released code (\texttt{pipeline/current/eval\_pilot/}). Long extractor prompts are shown as their core skeleton plus the hard-rule list, with the full text released with the code.

\subsection*{Alignment judge: shared base}
The same base prompt powers both \hall{}/\mis{}/\unc{} (report-to-source) and \rec{} (source-to-report); only the direction-specific context and rules differ.

\begin{tcolorbox}[
  enhanced, breakable, sharp corners,
  colback=anchGray, colframe=anchMuted, boxrule=0.4pt,
  left=4pt,right=4pt,top=4pt,bottom=4pt,
  fonttitle=\bfseries\small, title=Alignment judge (shared base)]
\small
You are given a \textsc{Claim} and a numbered list of \textsc{Evidence} statements (1 to N). The evidence was retrieved using semantic similarity and may include irrelevant, partially relevant, or redundant items. Each evidence line is prefixed with its cosine similarity score to the claim, higher scores indicate closer semantic match but do not guarantee factual support.

\{direction\_context\}

\textsc{Verdict definitions.}
\begin{itemize}[leftmargin=1.2em,itemsep=1pt,topsep=2pt]
\item \textsc{Supported}: One or more evidence statements clearly back the claim. The core factual content of the claim is present in the evidence, even if wording, structure, or level of detail differs. Minor differences in framing are acceptable as long as the central meaning is preserved.
\item \textsc{Partially Supported}: The evidence covers some but not all of the claim, or the match is incomplete or ambiguous (for example, confirms part of a compound statement but is silent on the rest, or provides weaker or less specific data than the claim asserts).
\item \textsc{Not Supported}: The evidence does not meaningfully back the claim. Use this when the claim's core assertion is absent, materially changed, or contradicted.
\end{itemize}

\textsc{Rules (shared).} (1) Use only the provided evidence text. Do not rely on outside knowledge. (2) Treat paraphrasing, summarisation, and synthesis as valid support when the factual meaning matches. (3) Ignore duplicate or redundant evidence lines, they do not strengthen support. \{direction\_rules\}

\textsc{Output (strict JSON, no markdown fences):}
\texttt{\{"verdict": "SUPPORTED | PARTIALLY SUPPORTED | NOT SUPPORTED", "confidence": <1-5>, "justification": "<5-20 word explanation>"\}}
\end{tcolorbox}

\subsection*{Direction: report-to-source (\hall{}, \mis{}, \unc{})}
\begin{tcolorbox}[
  enhanced, breakable, sharp corners,
  colback=anchGray, colframe=anchMuted, boxrule=0.4pt,
  left=4pt,right=4pt,top=4pt,bottom=4pt,
  fonttitle=\bfseries\small, title=Report-to-source direction]
\small
\textsc{Direction context.} The \textsc{Claim} is extracted from a generated report. The \textsc{Evidence} comes from original source documents. Your task: determine whether the source evidence factually supports the report's claim. Be strict, the report must be grounded in source data.

\textsc{Additional rules.} (4) Numeric precision matters: if the claim states a specific number and the evidence gives a materially different number, that part is not supported. (5) The claim may synthesise across sources, require that each key assertion has at least one evidence match.
\end{tcolorbox}

\subsection*{Direction: source-to-report (\rec{})}
\begin{tcolorbox}[
  enhanced, breakable, sharp corners,
  colback=anchGray, colframe=anchMuted, boxrule=0.4pt,
  left=4pt,right=4pt,top=4pt,bottom=4pt,
  fonttitle=\bfseries\small, title=Source-to-report direction]
\small
\textsc{Direction context.} The \textsc{Claim} is a fact extracted from an original source document. The \textsc{Evidence} consists of passages from a generated report. Your task: determine whether the report covers this source fact. The report may paraphrase, summarise, or incorporate the fact into broader analysis, this counts as coverage.

\textsc{Additional rules.} (4) The report may express the fact at a different granularity (e.g.\ rounded numbers, broader category). Accept this as coverage if the core information is preserved. (5) If the report subsumes the fact into a higher-level summary that clearly encompasses it, treat it as \textsc{Supported}.
\end{tcolorbox}

\subsection*{Relevance labeler (\textsc{High} / \textsc{Related} / \textsc{Unrelated})}
The relevance step has two implementations. The default ($\drwriter{}$ experiments) is a deterministic mapping from the extraction-time relevance field (\texttt{direct} becomes \textsc{High}, \texttt{off\_topic} becomes \textsc{Unrelated}, otherwise \textsc{Related}). The optional \emph{refine} path uses the following batched LLM relevance prompt.

\begin{tcolorbox}[
  enhanced, breakable, sharp corners,
  colback=anchGray, colframe=anchMuted, boxrule=0.4pt,
  left=4pt,right=4pt,top=4pt,bottom=4pt,
  fonttitle=\bfseries\small, title=Batched source-fact relevance labeler]
\small
You are labeling the relevance of factual sentences to a research task. For each fact, choose exactly one label: \textsc{High}, \textsc{Related}, or \textsc{Unrelated}.

\textsc{Task:} \{task\}

\textsc{Definitions and examples.}
\textsc{High}: directly answers the task, provides key evidence for the main question, or states a central statistic the report would need. \emph{Example} (task: ``What is Japan's elderly population trend?''): ``The proportion of people aged 65 and over in Japan reached 28.6\% in 2020 and is projected to peak around 2040.'' (label \textsc{High}).

\textsc{Related}: same topic or useful background but does not directly answer or strongly support the task. \emph{Example} (same task): ``The National Institute of Population and Social Security Research publishes official projections every five years.'' (label \textsc{Related}).

\textsc{Unrelated}: does not help answer the task; off-topic or different subject. \emph{Example} (same task): ``The company reported strong earnings in the retail segment last quarter.'' (label \textsc{Unrelated}).

\textsc{Facts to label} (each line is one fact; the leading number is the fact index, 0-based): \{numbered\_facts\}

\textsc{Output.} Exactly one line per fact index in the format ``index: LABEL''. Use only \textsc{High}, \textsc{Related}, or \textsc{Unrelated}. Example: \texttt{0: HIGH \textbackslash n 1: RELATED \textbackslash n 2: UNRELATED}. Output nothing else.
\end{tcolorbox}

\subsection*{Inconsistency judge (intra-report contradictions)}
\begin{tcolorbox}[
  enhanced, breakable, sharp corners,
  colback=anchGray, colframe=anchMuted, boxrule=0.4pt,
  left=4pt,right=4pt,top=4pt,bottom=4pt,
  fonttitle=\bfseries\small, title=Pairwise inconsistency judge]
\small
You are given \textsc{Claim\_A} and \textsc{Claim\_B}, both claims extracted from the same generated report. Decide whether they contradict each other.

\textsc{Verdict definitions.}
\begin{itemize}[leftmargin=1.2em,itemsep=1pt,topsep=2pt]
\item \textsc{Contradicts}: the two claims assert at least one specific fact about the same entity, quantity, time, and scope that cannot both be true (different numbers for the same measurement, opposite polarity, mutually exclusive categories, incompatible peak/trough years). A single concrete factual conflict is sufficient.
\item \textsc{Partially Contradicts}: the claims address overlapping subject matter and one tensions with the other without strictly conflicting (different granularity that could be reconciled, ambiguous scope, borderline rounding).
\item \textsc{Consistent}: both claims can be true together. They may be unrelated, orthogonal, complementary, or actually agreeing.
\end{itemize}

\textsc{Rules.} (1) Different time references about the same quantity are \textsc{Consistent}. (2) Different scopes (Japan vs Tokyo, total vs working-age) are \textsc{Consistent}. (3) Rounded vs precise numbers are \textsc{Consistent}. (4) Paraphrases of the same fact are \textsc{Consistent}. (5) Numeric disagreement on the same quantity, entity, time, and scope is \textsc{Contradicts}, even when the claims share other consistent content. (6) Use only the text given.

\textsc{Output.} Strict JSON: \texttt{\{"verdict": "CONTRADICTS | PARTIALLY\_CONTRADICTS | CONSISTENT", "confidence": <1-5>, "justification": "<5-20 word explanation>"\}}.
\end{tcolorbox}

\subsection*{Claim and fact extractor (shared core)}
Source-side fact extraction and report-side claim extraction share a common system prompt (\texttt{\_CLAIM\_PROMPT\_CORE}) of roughly four hundred lines; the report-side variant additionally requests inline citation reference IDs (\texttt{cited\_refs}). We reproduce the core's skeleton; the full text is released with the code.

\begin{tcolorbox}[
  enhanced, breakable, sharp corners,
  colback=anchGray, colframe=anchMuted, boxrule=0.4pt,
  left=4pt,right=4pt,top=4pt,bottom=4pt,
  fonttitle=\bfseries\small, title=Claim and fact extractor (skeleton)]
\small
A claim is a self-contained, specific, information-dense statement that can stand alone in a report without any surrounding context. Include all materially useful claims; if nothing is useful for the task, return an empty list.

\textsc{Language.} Output in English only, regardless of the language of the task description, report, or source content.

\textsc{Task-driven relevance.} Label each claim's relevance as one of \texttt{direct} (answers, proves, or quantifies a task question), \texttt{supporting} (evidence or proof point for a direct answer), \texttt{contextual} (useful background), or \texttt{off\_topic} (no meaningful connection). Score salience 0-10 (10 = answers a primary task question with hard data; 0 paired with off\_topic).

\textsc{Self-sufficiency.} Each claim must include \texttt{text} (fully standalone, no pronouns or generic placeholders, named subject inline), \texttt{time\_reference} (concrete calendar anchor only, never placeholder prose), \texttt{entities} (specific proper-noun named entities, no generic categories), \texttt{relevance}, and \texttt{salience\_score}.

\textsc{Hard rules.} No vague references (``it'', ``the company'', ``the study''); no relative time unless converted to a concrete year/date; no meta placeholders in \texttt{time\_reference} (e.g.\ ``report publication year'' without digits); no bare metrics without a named subject; no claim that depends on its heading or neighbours for meaning; preserve numeric precision; exclude boilerplate, navigation, and legal/privacy text.

\textsc{Quality gate.} Apply five checks to every claim before output: subject identifiable, event identifiable, time anchored (or correctly empty), scope present when available, claim serves a task section. If any fails, rewrite or drop.

\textsc{Output (JSON).} \texttt{\{"claims": [\{"text", "time\_reference", "entities", "relevance", "salience\_score"\}]\}}. Report-side adds \texttt{cited\_refs}: list of integer reference IDs found in or near the claim text.
\end{tcolorbox}

\subsection*{Fact-diff (Stage~1) source-change classifier}
This prompt is part of the writer's incremental-update path (Section~\ref{sec:writer}); the model receives the updated source page text and the prior facts extracted from that page restricted to the report-used candidates, and returns only facts whose value has changed as $(id, change)$ pairs where \texttt{change} is a concise natural-language instruction. The exact text lives with the writer release; the schema we evaluate against is reproduced below.

\begin{tcolorbox}[
  enhanced, breakable, sharp corners,
  colback=anchGray, colframe=anchMuted, boxrule=0.4pt,
  left=4pt,right=4pt,top=4pt,bottom=4pt,
  fonttitle=\bfseries\small, title=Source-change classifier (schema)]
\small
\textsc{Input.} Updated source page text (capped at the 200K-character extraction budget) and a listing of prior facts from that page, restricted to the candidates the report actually cites: \texttt{[\{id: fN, text: ...\}, ...]}.

\textsc{Task.} Return only the facts whose value the edit changed, each as an \texttt{(id, change)} pair, where \texttt{change} is a concise natural-language instruction describing the modification (for example, ``speech year changed from 1995 to 1998'') rather than a full rewritten fact.

\textsc{Output (JSON).} \texttt{\{"changes": [\{"id": "fN", "change": "..."\}]\}}. Emitting an instruction rather than a rewritten fact keeps Stage~1's completion-token cost near the floor.
\end{tcolorbox}

\end{document}